# Computer-Aided Automated Detection of Gene-Controlled Social Actions of Drosophilae

Faraz Ahmad Khan
Northumbria University
Newcastle upon Tyne, UK
faraz.khan@northumbria.ac.uk

Tiancheng Xia
Northumbria University
Newcastle upon Tyne, UK
tiancheng.xia@northumbria.ac.uk

Paul L Chazot
University of Durham,
Durham, UK
paul.chazot@durham.ac.uk

Abdel Ennaceur
University of Sunderland,
Sunderland, UK
abdel.ennaceur@gmail.com

Ahmed Bouridane
Northumbria University Newcastle upon Tyne, UK
ahmed.bouridane@northumbria.ac.uk

Richard Jiang
Northumbria University
Newcastle upon Tyne, UK
richard.jiang@northumbria.ac.uk

## ABSTRACT

Gene expression of social actions in Drosophilae has been attracting wide interest from biologists, medical scientists and psychologists. Gene-edited Drosophilae have been used as a test platform for experimental investigation. For example, Parkinson's genes can be embedded into a group of newly bred Drosophilae for research purpose. However, human observation of numerous tiny Drosophilae for a long term is an arduous work, and the dependence on human's acute perception is highly unreliable. As a result, an automated system of social action detection using machine learning has been highly demanded. In this study, we propose to automate the detection and classification of two innate aggressive actions demonstrated by Drosophilae. Robust keypoint detection is achieved using selective spatio-temporal interest points (sSTIP) which are then described using the 3D Scale Invariant Feature Transform (3D-SIFT) descriptors. Dimensionality reduction is performed using Spectral Regression Kernel Discriminant Analysis (SR-KDA) and classification is done using the nearest centre rule. The classification accuracy shown demonstrates the feasibility of the proposed system.

## KEYWORDS

Behaviour analysis, drosophila, automated action recognition, behavioural classification

## 1. INTRODUCTION

Social behaviour analysis has recently shown great promise in the field of computer vision and machine learning. Behavioural analysis has been used for annotating the actions of actors and players in videos [1, 2], has been immensely helpful in the modelling and development of sign languages [3] and has even been used in understanding the behaviour of animals [4].

Social behavioural analysis in drosophilae is an attractive option for researchers because analyzing humans in natural settings have technical as well as legal limitations. Legally, it is not possible to observe a human being at all times and therefore we miss on collecting spontaneous human activities. And technically it is very difficult to track and estimate a human's pose at all times, challenges such as uneven clothing, lighting and occlusions prevent the researchers from doing so. Furthermore, emulating and modelling the massive and complex architecture of a human brain requires weeks of processing time on a super computer – only a select few have access to such resources. Therefore it only makes sense to understand the thinking and working of a simpler brain that is similar to that of a human [5].

The common fruit fly (Drosophila Melanogaster) is an ideal model for studying neurobiology and social behaviour as researchers have observed huge similarities between the brain of a fruit fly and that of a mammal [6]. Furthermore, it has been observed that even with a nervous system that is far simpler than that of a mammal (135,000 neurons), a fruit fly is capable of performing complex behavioural actions [7]. From a genetic point of view it was observed that the drosophila genome is very much similar to that of a mammal [8]. These findings clearly suggest that by having a thorough understanding of the Drosophilae brain we can achieve a solid understanding of a mammalian brain.

Drosophilae have greatly helped in understanding neurodegenerative diseases such as Alzheimer's disease (AD) and Parkinson's disease (PD). This is possible because about 75% of disease causing genes found in humans can also be found in the fruit fly [9]. The drug discovery process is an expensive and slow procedure when using humans as test subjects but by modelling the disease onto the drosophilae, the process is greatly expedited due to the possibility of gene manipulation and having a large progeny. Researchers have acquired great insight into diseases such as PD and AD by reverse engineering the drosophilae genome sequence [10].

The innate social behaviour of aggression is of great interest in behavioural analysis and the drosophilae are one of the very few invertebrate genetic organisms that can demonstrate aggressive behaviour [11, 12]. Detecting and observing this behaviour is important as such stereotypical behaviour usually leads to a characteristic sequence [11]. In this paper, we propose to automate the detection of two aggressive social actions in the Caltech Fly-vs-Fly dataset that are similar to the naked eye and thus easily misclassified. We propose to classify between the hold action vs the tussle action. Hold (shown in Figure 1) is an aggressive action by the fly – where one fly holds onto the other after lunging. This action lasts for around 2500ms on average. The tussle (shown in Figure 2) is another aggressive action where the two flies lunge onto each other and repeatedly roll around while holding. This actions lasts for around 1170ms on average. It can be seen that from a machine learning point of view both of these actions look very similar and thus have the risk of being misclassified when observed manually.

We propose to use the selective spatio-temporal interest points (sSTIP) [13] for robust key point detection. sSTIP

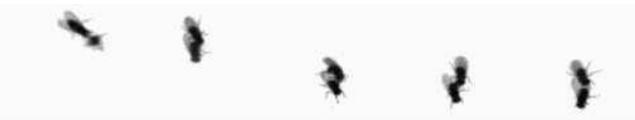

*Figure 1 Hold action between two fruit flies*

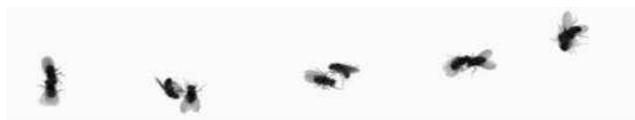

*Figure 2 A tussle action between two fruit flies*

have shown impressive performance in selecting only the most descriptive features when applied in human action recognition and was able to achieve very promising results on the KTH dataset [13].

3D-Scale invariant feature transform (3DSIFT) descriptors [14] are then extracted from the selected key points. SIFT has shown very promising results in object detection and recognition from images. 3D-SIFT applies the original SIFT descriptor on the video samples in 3D sub-volumes represented by its own sub-histogram. Once the features are extracted, they are used to generate a signature that can characterize each action of the fly. Kernel Discriminant Analysis using Spectral Regression (SR-KDA) [15] is used for dimensionality reduction and, finally, classification is performed using a nearest centre rule.

## 2. PROPOSED SYSTEM:

In this section, we present our proposed system of a binary classifier capable of identifying the above mentioned social behaviours in fruit flies.

**2.1 Selective Spatio-Temporal Interest Points:** The end goal of a strong feature detector is to be able to perform reliable behavioural analysis via automatic activity detection and action recognition. Laptev and Lindeberg [16] proposed spatio-temporal interest points (STIP) for effective action recognition. They proposed to extend the simple Harris corner detector [17] into spatio-temporal corners by detecting large intensity variations in both space and time. Recently, STIP based methods have gained popularity for use in action recognition applications. This is because methods based on STIP do not suffer from the temporal alignment problem and are invariant to viewpoint and geometric transformations [13]. However, despite their benefits, STIP-based methods do have their shortcomings: Some detected STIPs may be unstable and unreliable because of the local properties of the detector and there may be redundancy between descriptors extracted from adjacent STIPs.

Chakraborty et al. [13] proposed the sSTIP to address the shortcomings of the STIP detector. The authors proposed an improved version of STIP by supressing unwanted and redundant interest points and by imposing temporal and local constraints, thus achieving interest points that are more robust and less ambiguous.

First, spatial corners are detected using Harris corner detector $S_c$, where *c* represents the spatial scale. Then, an inhibition term for each detected $S_c$ is calculated for the purpose of surround suppression via a suppression mask. This is achieved by using a gradient weighting factor, defined as:

$$\delta_{\partial,\sigma}(a,b,a-y,b-z) \qquad (1)$$
$$= \cos(\partial_\sigma(a,b) - \partial_\sigma(a-y,b-z))$$

Where, $\partial_\sigma(a,b)$ and $\partial_\sigma(a-y,b-z)$ are the gradients at point $(a,b)$ and $(a-y,b-z)$, respectively. The horizontal and vertical range of the suppression mask are represented by $y$ and $z$, respectively.

A suppression term, $t_\sigma(a,b)$ that represents the weighted sum of gradient weights at each interest point, $S_c(a,b)$ is also defined:

$$t_\sigma(a,b) = \iint_\emptyset S_c(a-y,b-z) \qquad (2)$$
$$\times \delta_{\partial,\sigma}(a,b,a-y,b-y)dydz$$

Where $\emptyset$ represents the coordinate domain.

The detected corners $S_c$ and the suppression term $t_\sigma(a,b)$ are then fed to an operator, $C_{\rho,\sigma}$:

$$C_{\rho,\sigma} = H(S_c(a,b) - \rho t_\sigma(a,b) \qquad (3)$$

Where $\rho$ represents the suppression strength and $H(x)$ is given by:

$$H(x) = \begin{cases} x, & x \geq 0 \\ 0, & -x \end{cases} \qquad (4)$$

The corner magnitude $S_c(a,b)$ under evaluation is retained if no interest points are detected in the area surrounding it. However, detection of a large number of interest points in the surrounding area will result in a high suppression term $t_\sigma(a,b)$, thus causing the current corner point to be supressed. Constraints are further applied to the final set of suppressed corners $C_{\rho,\sigma}$ by using a non-maxima suppression algorithm, similar to that proposed in [18].

**2.2. 3D-SIFT Descriptors:** Now that the interest points have been detected, the next step is to describe the region around the interest point under evaluation using a spatio-temporal descriptor. 3D SIFT descriptors have been used in our work which is capable of describing the interest points that are robust to noise and orientation variations.

The authors in [14] proposed the 3D-SIFT descriptors by extending the popular 2D-SIFT descriptor into spatio-temporal dimensions. In 3D, the magnitude and orientations for each pixel is given by the following equations:

$$M_{3D}(a,b,t) = \sqrt{L_a^2 + L_b^2 + L_t^2} \qquad (5)$$

$$\theta(a,b,t) = \tan^{-1}(L_b/L_a) \qquad (6)$$

$$\emptyset(a,b,t) = \tan^{-1}\left(L_t / \sqrt{L_a^2 + L_b^2}\right) \qquad (7)$$

Where $M_{3D}$ represents the magnitude in 3D, $\theta$ represents the angle and $\emptyset$ represents the angle that is encoded away from the 2D gradient. $L_a$ and $L_b$ are determined using finite difference approximations.

Once each point can be represented using the above three equations, the next step is to generate a weighted histogram of the region surrounding the interest point. This is accomplished by dividing the angles ($\theta$ and $\emptyset$) into bins of equal size. Each bin is then normalized using a solid angle $\sigma$, which is calculated as:

$$\sigma = \Delta\emptyset(\cos\theta - \cos(\theta + \Delta\theta)) \qquad (8)$$

The SIFT descriptor sub-histograms are extracted from sub-regions of ($n \times n \times n$) that surround the interest point. The value that is added to the sub-histograms are given as:

$$hist(i_\theta, i_\emptyset) += \frac{\frac{1}{\sigma}M_{3D}(a',b',t') \times}{e^{\frac{-((a-a')^2+(b-b')^2+(t-t')^2}{2\sigma^2}}} \qquad (9)$$

Where $(a,b,t)$ represents the location of the current pixel and $(a',b',t')$ represents the location of the interest point that is to be added to the histogram.

**2.3. Spectral Regression Kernel Discriminant Analysis (SR-KDA):** Kernel Discriminant Analysis (KDA) is the non-linear version of Linear Discriminant Analysis (LDA). Non-linear adaptation of LDA can be performed using a $n \times n$ kernel matrix, $K$ that is generated from the training data. The KDA objective function is given by:

$$max_\tau D(\tau) = \tau^T M_b \tau / \tau^T M_t \tau \qquad (10)$$

Where $M_b$ and $M_t$ represent the between class scatter matrix and total feature space scatter matrix, respectively and $\tau$ represents the projection function into the kernel space. Equation (10) can also be written as [19]:

$$max_\omega D(\omega) = {\omega^T KLK\omega}/{\omega^T KK\omega} \qquad (11)$$

Where $\omega = [\omega_1, \omega_2, \omega_3, \cdots \omega_n]^T$ is an eigenvector where every $\omega$ gives a projection of $\tau$ into the feature space and $L$ is a diagonal block matrix of action labels.

It was further discovered in [20], that rather than focusing on the KDA eigenproblem, the following two equations can also be used to calculate KDA projections:

$$L\varpi = \lambda\varpi \qquad (12)$$

$$(K + \delta I)\omega = \varpi \qquad (13)$$

Where $\varpi$ is an eigenvector of $L$, $I$ is the identity matrix and the regularization parameter is represented by $\delta > 0$.

## 3. RESULTS AND DISCUSSION:

As mentioned in Section 1, key point detection has been performed using sSTIP, which provides a robust detection of key points. sSTIP was shown to perform very well when applied on the KTH dataset by discarding redundant features and targeting only the most descriptive features for human action recognition [13].

These selected key points are described using the 3D-SIFT descriptor. SIFT has consistently shown to produce good results in matters related to object matching and detection in images. 3D-SIFT is an extension of the original SIFT descriptor which allows it to be applied on video samples, this is achieved by representing 3D sub volumes by their respective sub histograms (See Section 2.2). These features are then fed to SR-KDA for dimensionality reduction and finally the system is ready for classification of actions.

The Caltech Fly-vs-Fly dataset [21] is used to test the performance of our proposed system. The dataset is composed of 47 pairs of flies that demonstrate 10 different social behaviours. Three different settings are used for observing fly engagement: aggression, courtship and boy meets boy. Experts have annotated all behaviours observed in the recorded videos. Figure 3 shows the experimental setup for each setting.

From the dataset, only the videos demonstrating the hold and tussle actions are selected. The snippets of video demonstrating each action are extracted using the provided annotations. 50 samples of each action are acquired, out of which 35 are used for training while the rest are used for

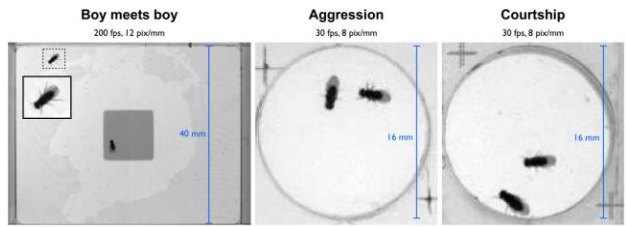

*Figure 3. Boy meets boy are two male flies in a large chamber containing a food patch recorded at very high resolution. Aggression and Courtship videos are recorded from comparatively smaller chambers having uniform food surfaces and are recorded at lower resolutions.*

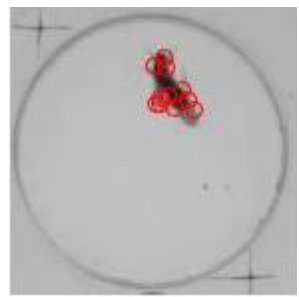

*Figure 4. Detection of sSTIP from a single frame of video*

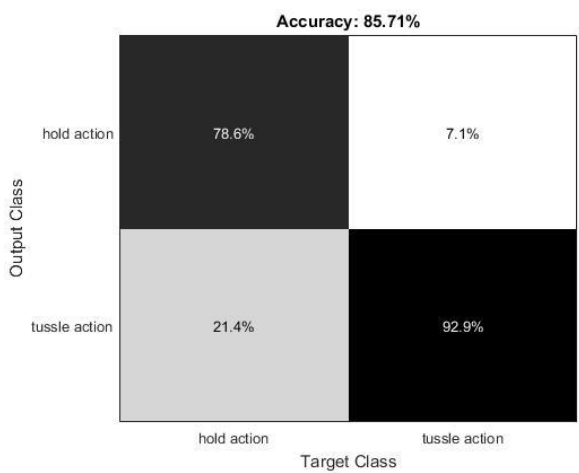

*Figure 5. Confusion matrix displaying the classification results*

testing.

sSTIP are detected from every video using the following settings: block dimension of 3, alpha value of 1.5 and a temporal scale of 5. Figure 4 demonstrated the sSTIP detection process. A 640 dimensional 3D-SIFT descriptor is then calculated at every detected key point. Once the 3D-SIFT descriptors from all the training videos have been extracted, they undergo dimensionality reduction using SR-KDA; this step provides us with a model which can be used for classifying an unknown action. The testing videos also go through the same process. Using the above mentioned settings, we were able to correctly predict the known videos with an accuracy of 85.71%. The confusion matrix of the classification results is shown in Figure 5. As can be seen the system is capable of classifying the tussle actions with more confidence. This can be explained by the fact that the tussle videos are more descriptive in terms of feature points and a lot more is going on in the videos when compared to the videos showing the hold action. This results in more descriptive features for the tussle action and, thus, superior results in classifying this action as well.

## 4. CONCLUSIONS

Social behaviour analysis has been of great interest in the field of computer vision and machine learning. Understanding the instinctive behaviours demonstrated by the Drosophilae or gene-edited Drosophilae is of great interest to neurobiologists as by doing so we can greatly advance our understanding of the mammalian brain. In this paper we propose to classify two instinctive and characteristic behaviours of the Drosophilae: the hold action and the tussle action. Robust keypoints were detected using sSTIP and described using 3D-SIFT. SR-KDA was used for dimensionality reduction and classification showed an accuracy of 85.71% which demonstrates the feasibility of the proposed system.

## ACKNOWLEDGMENTS

This work is supported by EPSRC grant (EP/P009727/1). The authors declare that there is no conflict of interest regarding the publication of this paper.